\documentclass[10pt]{article}
\usepackage[linesnumbered,ruled,vlined]{algorithm2e}

\SetCommentSty{mycommfont}

\usepackage{multirow}
\usepackage{amsfonts} 
\usepackage[letterpaper]{geometry}
\usepackage{hicss51}
\usepackage{times}
\usepackage[none]{hyphenat}
\usepackage{url}
\usepackage{latexsym}
\usepackage{hyperref}
\usepackage{breakurl}
\hypersetup{
    colorlinks = true,
    linkcolor = red,
    anchorcolor = blue,
    citecolor = blue,
    filecolor = blue,
    urlcolor = blue
}
\usepackage[english]{babel}
\usepackage{apacite}
\usepackage{comment}
\usepackage{url}
\usepackage{latexsym}

\usepackage{graphicx}

\usepackage{array}

\usepackage{listings}
\usepackage{tabularx}
\usepackage{booktabs}
\usepackage{threeparttable}
\usepackage{multirow}
\usepackage{tabularx}
\usepackage{bigstrut}
\usepackage{amsmath}
\usepackage{amsfonts}
\usepackage{amssymb}
\usepackage{wasysym}
\usepackage{enumitem}

\usepackage{xcolor}

\usepackage{xurl}
\usepackage{hyperref}
\SetKwComment{Comment}{$\triangleright$\ }{}

\title{Differentially private fine-tuned NF-Net to predict GI cancer type}

 \author{Sai Venkatesh Chilukoti \\
  University of Louisiana at Lafayette \\
  {\underline{sai-venkatesh.chilukoti1@louisiana.edu} }\\ \\
  Vijay Srinivas Tida\\
 College of Saint Benedict and Saint John’s University \\
  {\underline{VTIDA001@CSBSJU.EDU}} \\ \And
  Md Imran Hossen \\
  University of Louisiana at Lafayette\\
  {\underline{md-imran.hossen1@louisiana.edu} } \\ \\
\\ \And
  Liqun Shan\\
 University of Louisiana at Lafayette\\
 {\underline{liqun.shan1@louisiana.edu} } \\ \\ 
   Xiali Hei\\
 University of Louisiana at Lafayette\\
  {\underline{xiali.hei@louisiana.edu }} \\ \\}

\date{}

\begin{document}
\maketitle
\begin{abstract}
 
 Based on global genomic status, the cancer tumor is classified as Microsatellite Instable (MSI) and Microsatellite Stable (MSS). Immunotherapy is used to diagnose MSI, whereas radiation and chemotherapy are used for MSS. Therefore, it is significant to classify a gastro-intestinal (GI) cancer tumor into MSI vs. MSS to provide appropriate treatment. The existing literature showed that deep learning could directly predict the class of GI cancer tumors from histological images. However, deep learning (DL) models are susceptible to various threats, including membership inference attacks, model extraction attacks, etc. These attacks render the use of DL models impractical in real-world scenarios. To make the DL models useful and maintain privacy, we integrate differential privacy (DP) with DL. In particular, this paper aims to predict the state of GI cancer while preserving the privacy of sensitive data. We fine-tuned the Normalizer Free Net (NF-Net) model. We obtained an accuracy of 88.98\%  without DP to predict (GI) cancer status. When we fine-tuned the NF-Net using DP-AdamW and adaptive DP-AdamW, we got accuracies of 74.58\% and 76.48\%, respectively. Moreover, we investigate the Weighted Random Sampler (WRS) and Class weighting (CW) to solve the data imbalance. We also evaluated and analyzed the DP algorithms in different settings.

\end{abstract}

\subsubsection*{Keywords:}

 Deep Learning, DP-AdamW, GI Cancer, NF-Net, and adaptive DP-AdamW.

\section{Introduction}

The International Agency for Research on Cancer (IARC) \cite{IARC} estimates that GI cancer causes 35\% of global cancer-related deaths. Based on their global genomic status, GI cancers can be classified as MSI and MSS. MSI tumors are more likely to respond to immunotherapy, whereas MSS tumors respond to chemotherapy and radiation. Therefore, it is important to be able to distinguish between MSI and MSS tumors. Deep convolutional neural networks (DCNNs)~\cite{kather2019deep}, such as ResNets~\cite{he2016deep}, have been shown to be effective in classifying MSI and MSS tumors~\cite{sai2022modified, yamashita2021deep, khan2022transfer}. However, they require a lot of data ~\cite{DL} to train and can leak sensitive information. DL models are susceptible to attacks such as membership inference attacks~\cite{7958568}, model extraction attacks~\cite{MLSecurity}, and model inversion attacks~\cite{fredrikson2014privacy}. 

Therefore, the development of predictive models that protect patient privacy is important ~\cite{HIPAA,P&S,Issues}. The Health Insurance Portability and Accountability Act of 1996 (HIPAA) establishes rules and regulations to maintain patient privacy. It establishes rules and regulations for how healthcare providers collect, use, and share patient information. Deep learning models that are used to classify MSI and MSS from histological images must adhere to HIPAA to protect patient privacy. There are a number of ways to design deep learning models that adhere to HIPAA. One approach is to use DP. DP is a technique that adds noise to data to protect individual privacy. 

Most of the prior work on DL models~\cite{he2016deep,simonyan2014very} ~\cite{chollet2017xception} for MSI vs. MSS classification of GI cancer did not consider privacy. These models were trained on large datasets of images, which could be a privacy risk. To provide security to the DL model against the above-described attacks, we implement a DP ~\cite{DP} based DL model. More specifically, we use the DP-AdamW and adaptive DP-AdamW mechanisms to fine-tune the NF-Net model~\cite{brock2021high} for MSI vs. MSS classification of GI Cancer. 

Our main contributions to this project are as follows:

\begin{itemize}
   \item We propose a model for classifying MSI vs. MSS using an NF-Net-F0 pre-trained feature extractor combined with a custom classifier. Then, we evaluated the model on a dataset~\cite{Dataset} and got an accuracy of 76.48\% when trained using DP-AdamW. In contrast, the non-private model obtains an accuracy of 89.81\%.
   \item We extensively evaluate the privacy-enhanced model in different settings, including different noise scales, different optimizers, different epsilon, etc. Then, compare our work with the existing literature. 
   \item We also investigate the techniques to deal with data imbalance, like a WRS and CW under DP.
\end{itemize}

The subsequent sections of the paper are structured as follows: Section 2 provides an overview of the background information, including DP and its noise-adding mechanisms. It further delves into the concepts of R\'enyi DP and truncated concentrated DP privacy accountants, along with a literature review. Section 3 presents the methodology of our work. Section 4 outlines the evaluation of our results. Section 5 explores the limitations and potential future directions of our study. Finally, in Section 7, we provide a conclusion to our work.

\section{Background and literature review}

Deep learning models have been successfully used in various applications~\cite{tida2022universal, tida2022unified}, including medical applications primarily related to image-based diagnosis \cite{lundervold2019overview}. For example, deep-learning models have been used to classify diseases such as lung disease, diabetic retinopathy~\cite{chilukoti2023reliable}, malaria, and various cancers\cite{kieu2020survey}. However, using deep learning models in the medical field raises privacy concerns. For example, deep learning models can be used to infer sensitive information from medical images, such as the patient's identity, medical history, or diagnosis. To address these privacy concerns, researchers have developed privacy-preserving deep learning models for applications such as Covid-19~\cite {chilukoti2022privacy}.                  

\subsection{Differential Privacy and its noise adding mechanisms}

While conducting data analysis, DP is employed to safeguard the sensitive information of individual data points and only disclose aggregated data results. Developing DP algorithms involves incorporating noise-adding mechanisms like Laplacian and Gaussian distributions~\cite{dwork2006calibrating,dwork2014algorithmic, DP2}. The level of effectiveness of DP is measured using the parameters $\epsilon$ and $\delta$~\cite{DP3}. $\epsilon$ quantifies the extent to which the output of a model changes when a single data point is included or excluded from a dataset. On the other hand, $\delta$ represents the probability of unintentionally revealing information about specific data samples. If both $\epsilon$ and $\delta$ are small, the privacy of individuals is higher, but it comes at the cost of reduced accuracy. Conversely, larger values of $\epsilon$ and $\delta$ correspond to lower privacy and higher accuracy. DP can be defined as a randomized algorithm $M: D-> R$, where $D$ represents the domain and $R$ denotes the range. The algorithm $M$ is considered ($\epsilon$,$\delta$)-differentially private if, for any subset of outputs $S \subseteq R$ and any two neighboring inputs $d$, $d'\in D$, the following inequality holds: $Pr[M(d) \in S]$ $\leq$ $e^\epsilon$ $Pr[M(d')\in S] + \delta $. 

The primary objective of DP is to defend the privacy of individuals while still being able to derive general insights. For instance, in the realm of DL, adding or removing a single data sample should not significantly impact the classifier's predictions. The researcher's pivotal purpose is to find a balance that allows the maximum extraction of the necessary information to perform the given task without compromising individual privacy \cite{dwork2008differential}. DP has become increasingly crucial in various domains, including government agencies, the healthcare sector\cite{ficek2021differential}, and service providers \cite{wang2020edge,gai2019differential}.

In the domain of DL, the Laplacian and Gaussian noise mechanisms are two widely utilized \cite{dwork2014algorithmic,jain2018differential} and popular methods for introducing noise. The Gaussian Mechanism (GM) offers stronger privacy protection than the Laplacian mechanism. This is due to the fact that GM introduces less noise into the model because of $l_2$ sensitivity. While GM does not satisfy pure $\epsilon$-DP, it does satisfy ($\epsilon$, $\delta$)-DP. 
 
\subsection{R{\'e}nyi Differential Privacy}

Rényi DP (RDP) has emerged as a novel extension of $\epsilon$-DP, introduced in ~\cite{mironov2017renyi}. It provides a comparable alternative to the ($\epsilon, \delta$) formulation, making it easier to interpret. RDP is considered a natural relaxation of DP, utilizing the concept of Rényi divergence. When considering random variables $X$ and $Y$, which can assume $n$ possible values with positive probabilities $p_i$ and $q_i$, respectively, the Rényi divergence of $X$ from $Y$ can be defined using the following equation~\ref{eq01}:  

\begin{equation}\label{eq01} 
\begin{array}{l}
    D_{\alpha}(X||Y)= \frac{1}{(\alpha-1)}log(\sum\limits_{i=1}^n \frac{p_{i}^{\alpha}}{q_{i}^{\alpha-1}})
 \end{array}    
\end{equation}

Where $\alpha > 0$ and not equal to 1. However, special cases will be obtained when $\alpha =0, 1, \infty$. As $\alpha$ converges to 1, $D_{\alpha}$ converges to the Kullback-Leibler (KL) divergence \cite{joyce2011kullback}. RDP has nice composition properties that help compute the model's total privacy budget. RDP privacy falls somewhere in between $\epsilon$-DP and ($\epsilon, \delta$)-DP. 

\subsection{truncated Concentrated Differential Privacy}

If, for all values of $\tau$ within the range of $(1, \omega)$, a randomized algorithm $\mathcal{A}$ satisfies the condition of being $(\rho, \omega)$-truncated concentrated DP, then for any pair of neighboring datasets $d$ and $\hat{d}$, and any value of $\alpha$ greater than 1, the following inequality holds. 

\begin{equation}
     D_{\tau}(\mathcal{A}(d)||\mathcal{A}(\hat{d})) \leq \rho\alpha
\end{equation}

where $D_{\tau}(.||.)$ is the Renyi divergence of order $\tau$. 

Given two distributions $\mu$ and $\nu$ on a Banach space $(Z,|| . ||)$, Rényi divergence is computed as follows. 

Rényi Divergence~\cite{renyi1961measures}: Let $1 < \alpha < \infty$,  and $\mu, \nu$ be measures with $\mu \ll \nu$. The Rényi divergence of orders $\alpha$ between $\mu$ and $\nu$ between is defined as:  

\begin{equation}
        D_{\tau}(\mu||\nu) \doteq \frac{1}{\alpha - 1}ln\int(\frac{\mu(z)}{\nu(z))})^{\alpha}\nu(z)dz.
\end{equation}

We consider $\frac{0}{0}$ to be equal to 0. If $\mu$ is not absolutely continuous with respect to $\nu$, we define the Rényi divergence as infinity. The Rényi divergence for orders $\alpha = 1$ and $\alpha = \infty$ is defined based on continuity.

\subsection{Literature review}

In their study, ~\cite{puliga2021microsatellite} demonstrated the positive impacts of immunotherapy in treating patients with GI cancer who have MSI status. Their research emphasizes the importance of accurately classifying patients into MSI and MSS categories in order to identify those who could benefit from specific treatment approaches, particularly immunotherapy. In a thorough investigation by ~\cite{zhao2022identification}, various tasks related to gastric cancer, including classification and segmentation, were extensively studied. In their research, ~\cite{li2019early} employed fluorescence hyperspectral imaging to capture fluorescence spectral images. They integrated deep learning into spectral-spatial classification methods to effectively identify and extract combined ``spectral + spatial" features. These features were utilized to construct an early diagnosis model for GI cancer. The model achieved an overall accuracy of 96.5\% when classifying non-precancerous lesions, precancerous lesions, and gastric cancer groups.

~\cite{sai2022modified} introduced a modified ResNet model for classifying MSI vs. MSS in GI cancer. Their proposed model achieved a test accuracy of 89.81\%. In their study, ~\cite{khan2022transfer} utilized a transfer learning-based approach to construct a model for classifying MSI vs. MSS for cancers. The model was trained on histological images obtained from formalin-fixed paraffin-embedded (FFPE) samples. The results demonstrated an accuracy of 90.91\%. In their work, ~\cite {chilukoti2022privacy} described the threats related to medical applications and developed a DP-friendly model for Covid19 detection to nullify the threats posed to the non-private model.

Several important observations can be made based on the review of existing literature. First, classifying MSI vs. MSS in GI cancer is crucial in ensuring patients receive appropriate treatment. It is important to acknowledge that deep learning models are susceptible to various attacks. Interestingly, prior research in the context of GI cancer classification has often overlooked privacy-related concerns during the training of classifiers. Drawing inspiration from the medical applications of DP, we develop a model incorporating DP and leveraging a privacy-friendly pre-trained model known as NF-Net. 

\section{Methodology}


\subsection{Rationale for NF-Net model}

We select the NF-Net model because of the ability to replicate the benefits offered by Batch Normalization (BN) using scaled weight standardization and adaptive gradient clipping etc. Moreover, NF-Nets provide better accuracy than the BN-based models such as ResNet~\cite{he2016deep}. BN is a widely employed method in deep learning that enhances the training process of neural networks. It standardizes the activations within each layer by subtracting the mean and dividing by the standard deviation of the batch. This normalization technique effectively reduces the impact of internal co-variate shifts, enhances the flow of gradients, and expedites the convergence of the model. When considering DP, particularly in the context of DP-SGD, there are difficulties in directly utilizing BN. This is primarily due to the fact that BN introduces inter-dependencies among individual examples within a batch. Such inter-dependencies can potentially compromise the privacy offered by DP. 

To overcome this challenge, researchers have proposed alternative methods, such as layer normalization and instance normalization, that strive to achieve privacy-preserving normalization while upholding DP guarantees. ~\cite{brock2021high} proposed (NF-Nets) that are shown to be more efficient than BN-based networks like ResNet and VGG. Thus, we choose the NF-Net as the pre-trained feature extractor to implement DP-SGD and adaptive DP-SGD methods on GI cancer status prediction. Moreover, scaled weighted standardization blocks are DP-friendly as they do not involve batch statistics. 

\subsection{Overview of differentially private optimizers}

\begin{figure}[ht]
    \centering
     \begin{small}
     \includegraphics[scale=0.4, keepaspectratio]{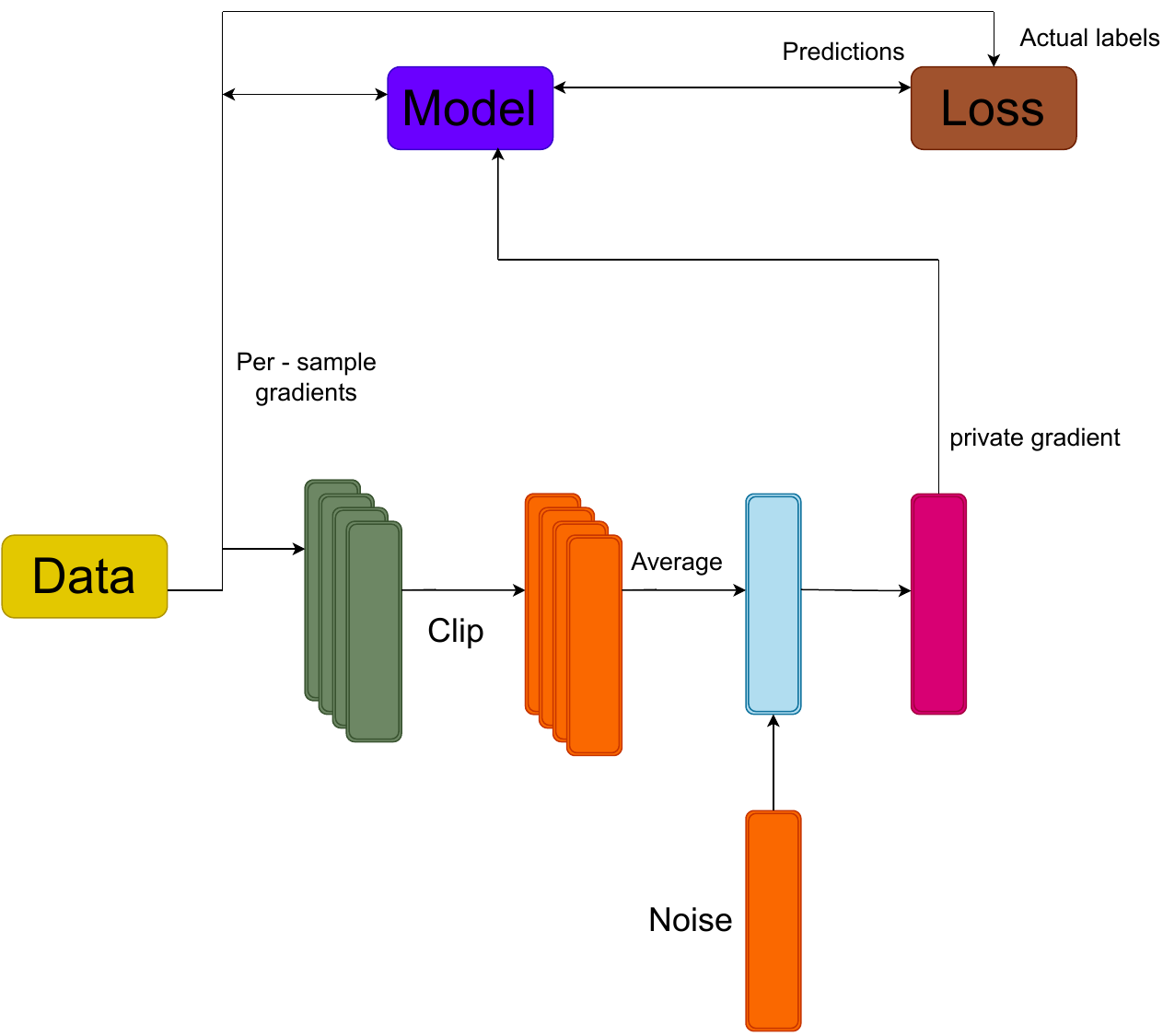}
    \caption{\textbf{Illustration of DP-SGD and its variants}}
    \label{DP}
    \vspace{-3mm}
    \end{small} 
    \end{figure}

The procedures of DP-SGD, DP-AdamW, and Adaptive DP-AdamW are similar, with minor variations. Hence, we will first discuss the general process common to all these techniques. Figure ~\ref{DP} provides an overview of the end-to-end process of DP-SGD and its variants. The process begins with the model receiving input data and making predictions based on that data. The predictions and the actual labels are then passed to the loss function, which calculates the loss value. Using this loss value, gradients are computed for each model parameter with respect to each sample in the batch. These per-sample gradients are clipped to control their magnitude and then averaged, and noise is added to generate a private gradient. Finally, the model is updated using the private gradients, and this iterative process continues until convergence is achieved. The key distinction between DP-SGD and its variants, as compared to traditional SGD (Stochastic Gradient Descent) and its variants, lies in the per-sample gradient clipping and addition of noise to compute the average gradient. 

\subsection{Differentially Private AdamW optimizer and its adaptive variant} 
\RestyleAlgo{ruled}
\SetKwComment{Comment}{/* }{ */}

\begin{algorithm}[h!]
\caption{Adaptive Differentially Private AdamW (ADP-AdamW)}
\SetKwInput{KwData}{Input}
\SetKwInput{KwResult}{Output}
\label{alg:dpsgd}
\KwData{Examples \{$x_1, ...,x_N$\}, loss function $\mathcal{L}(\theta) = \frac{1}{N}\sum_i\mathcal{L}(\theta, x_i)$. Parameters: learning rate $\eta_t$, noise scale $\sigma_t$, group size $L$, gradient norm bound $S$, $\alpha=0.001$, $\beta_{1} = 0.9$, $\beta_{2} = 0.999$, $\epsilon = 10^{-8}$, noise decay $R = 0.99, \lambda \in \mathbb{R}.$}
\textbf{Initialize} parameter vector $\theta_0$ randomly, time step $t =0$, first moment vector $m_{t=0} \leftarrow 0$, second moment vector $v_{t=0} \leftarrow 0$, schedule multiplier $\eta_{t=0} \in \mathbb{R}$, noise scale $\sigma_{t} \in \mathbb{R}^{+}$ \\
\KwResult{$\theta_T$ and compute the overall privacy budget $(\epsilon, \delta)$ using the RDP/tCDP privacy accountant.}
\For{$t = 0, . . . , T-1$}{
    Take a random sample $L_t$ with sampling probability $L/N$ \\
    For each $i \in L_t$, compute $\textbf{g}_t(x_i) \gets \Delta_{\theta_{t}}\mathcal{L}(\theta_t, x_i)$    \\
    $\bar{\textbf{g}}_t(x_i) \gets \textbf{g}_t(x_i)/max(1, \frac{||\textbf{g}_t(x_i)||_2}{S})$,    \\
    $\tilde{\textbf{g}}_t \gets \frac{1}{L}(\sum_i(\bar{\textbf{g}}_t(x_i) + \mathcal{N}(0, \sigma_t^2S^2\mathbf{I}))$,\\
    $\sigma_{t+1}^2 = R\sigma_t^{2}$,\\
    $\textbf{m}_{t} \gets  \beta_{1}\textbf{m}_{t-1} + (1 - \beta_{1})\tilde{\textbf{g}}_t$,\\
    $\textbf{v}_{t} \gets \beta_{2}\textbf{v}_{t-1} +(1 - \beta_{2})\tilde{\textbf{g}}_t^2$,\\
    $\hat{\textbf{m}}_t \gets \textbf{m}_t/(1-\beta_{1}^t)$,\\
    $\hat{\textbf{v}}_t \gets \textbf{v}_t/(1-\beta_{2}^t)$,\\
    $\eta_{t} \gets SetScheduleMultiplier(t)$\\
    $\theta_{t+1} \gets \theta_t - \eta_{t}(\alpha\hat{\textbf{m}}_t/(\sqrt{\hat{\textbf{v}}_t} + \epsilon) + \lambda\theta_{t-1})$\\    
} 
\end{algorithm}

This section focuses on the DP-AdamW~\cite{abadi2016deep} algorithm and its adaptive variant~\cite{zhang2021adaptive}, which we use for training models. The two algorithms differ primarily in computing the noise scale during each iteration. The training iteration of both DP-AdamW and its adaptive variant begins by randomly sampling a batch of training examples. The model predicts the outputs for each sample, and the loss is computed based on these predictions. Then, we estimate the gradients for each sample and average these gradients. Next, we add noise to the averaged gradient, and if it is the adaptive variant, we compute the noise scale for the next iteration accordingly. In the regular DP-AdamW, the noise scale remains the same throughout the training process. Then, we compute the first-moment, second-moment vectors, and corresponding bias-corrected moment vectors. Next, the learning rate is determined using a learning rate scheduler, and then gradient descent is applied to update the model parameters. This entire process is repeated for all the training iterations. Finally, if DP-AdamW is used, the privacy budget is computed using an RDP accountant. Alternatively, if it is the adaptive variant, tCDP determines the privacy guarantees. 

\subsection{Techniques for data imbalance}

In this section, we will explore two approaches employed in our study to address the issue of data imbalance: the WRS and CW. The WRS assigns weights to individual samples in a dataset, with samples from classes with more data receiving lower weights and vice versa. During training, the WRS prioritizes the selection of samples with higher weights, resulting in more frequent selection, while samples with lower weights are chosen less frequently. This strategy effectively balances the data distribution within the training process, mitigating the impact of data imbalance. 

In contrast, the CW technique assigns weights to each class instead of individual samples. It entails assigning a higher weight to the minority class and a lower weight to the majority class. These class weights are then incorporated into the loss function during training. By doing so, the loss function imposes greater penalties for misclassifying samples from the minority class. This strategy encourages the model to prioritize the minority class and enhances its capacity to learn from imbalanced data.  

\section{Results}


\subsection{Dataset description and hyper-parameter details}

The dataset~\cite{Dataset} we use in this project consists of a total of 192,000 histological images. The dataset has been divided into three subsets: 10\% for testing, 80\% for training, and another 10\% for validation. The training images were resized to 192$\times$192, while the test and validation images were resized to 256$\times$256. For further details regarding the dataset used in the project, refer to Table \ref{32}.

\begin{table}[!hbtp]
\centering
\caption{Dataset Description}
\begin{tabular}{|l|l|l|}
\hline
\textbf{Dataset} & \textbf{MSI}   & \textbf{MSS}   \\ \hline
Train   & 60,031 & 93,818 \\ \hline
Test    & 7,505  & 11,728 \\ \hline
Valid   & 7,503  & 11,727 \\ \hline
\end{tabular}

\label{32}
\end{table}

During our experiments, we employed a batch size of 64 and set the learning rate to $1e-4$. We utilized a one-cycle learning rate scheduler and the AdamW optimizer. The clipping value was set to 10, and no oversampling was performed. We set the epsilon value to 1. Throughout the experiments, we varied these hyper-parameters to observe their impact on the results. In all the below tables, we denote the DP-AdamW as DP and Adaptive DP-AdamW as ADP.

\subsection{Details of privacy computation}  
The privacy computation for DP-AdamW is performed using RDP accountant from ~\cite{TF-Privacy}. We use tCDP accountant to estimate the privacy loss for Adaptive DP-AdamW. We use the following expressions to find the total privacy budget using tCDP.

\begin{equation}
    \centering
    \label{rho,omega}
    \rho, \omega = (\frac{13(B/M)^2 C^2 (1 - R^E)}{2\sigma_0^2(R^{E-1} - R^E)} , \frac{\log(M/B)\sigma_0^2 R^{E-1}}{2C^2})
\end{equation}

$B$ is the batch size, $C$ is the clip, $R$ is the decay factor, $E$ is the epoch number, $\sigma_0$ is the initial noise multiplier, $M$ is the total dataset.   

Next, we convert the $\rho$ to $\epsilon$ using equation~\ref{eps} after setting the delta to be 1e-5.

\begin{equation}
    \label{eps}
    \centering
    \epsilon = (\rho +2\sqrt{\rho ln(1/\delta)})
\end{equation}

\begin{figure}[!hbtp]
    \centering
     \includegraphics[scale=0.65]{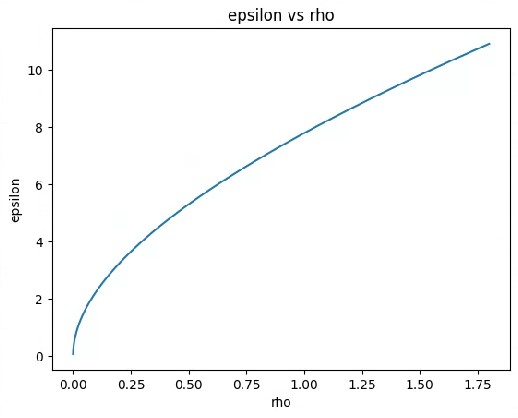}
    \caption{\textbf{Rho ($\rho$) versus epsilon ($\epsilon$)}}
    \label{RE}
    \vspace{-3mm}
    \end{figure}

We use the five different $\epsilon$ : (1,2,5,8, and 10) to compare DP-AdamW and Adaptive DP-AdamW. For the RDP accountant, the noise for a given $\epsilon$ is computed using the TF-Privacy package. Whereas, for tCDP, first, we find the $\rho$ for given $\epsilon$ using the plot~\ref{RE}. Next, we use this $\rho$ to compute the $\sigma_0$ using equation~\ref{rho,omega}.

\subsection{Epsilon vs. Metric scores}

We evaluate the model's performance across different epsilon values (1, 2, 5, 8, and 10) by analyzing the metric scores. Table ~\ref{epsilons_T} presents the scores for both DP and ADP. ADP consistently outperforms DP in all the metrics. For example, with an epsilon of 10, DP achieves an accuracy of 75.50\%, precision of 75.15\%, recall of 89.22\%, F1-score of 81.66\%, and ROC score of 71.60\%, while ADP achieves higher scores of 76.65\%, 77.40\%, 90.35\%, 81.99\%, and 73.70\% respectively. Similar trends are observed for other epsilon values, indicating the superior performance of ADP over DP. Notably, as epsilon increases, the metric scores also increase, suggesting a trade-off between privacy and utility. For instance, in Table ~\ref{epsilons_T}, the accuracy scores for DP are 74.58\%, 94.95\%, 75.25\%, 75.42\%, and 75.50\% for epsilon values of 1, 2, 5, 8, and 10 respectively. In comparison, ADP achieves accuracy scores of 75.78\%, 76.35\%, 76.48\%, 76.57\%, and 76.65\%. This pattern is consistent for other metrics as well.        

\begin{table*}[!hbtp]
\centering
\caption{Epsilon versus metric scores for DP-Adam and Adaptive DP-AdamW}
\begin{tabular}{|l|ll|ll|ll|ll|ll|}
\hline
$\epsilon$ & \multicolumn{2}{l|}{Accuracy}          & \multicolumn{2}{l|}{Precision}         & \multicolumn{2}{l|}{Recall}            & \multicolumn{2}{l|}{F1 score}          & \multicolumn{2}{l|}{ROC score}         \\ \hline
           & \multicolumn{1}{l|}{DP}      & ADP     & \multicolumn{1}{l|}{DP}      & ADP     & \multicolumn{1}{l|}{DP}      & ADP     & \multicolumn{1}{l|}{DP}      & ADP     & \multicolumn{1}{l|}{DP}      & ADP     \\ \hline
1          & \multicolumn{1}{l|}{74.58\%} & 75.78\% & \multicolumn{1}{l|}{74.12\%} & 75.50\% & \multicolumn{1}{l|}{87.01\%} & 89.37\% & \multicolumn{1}{l|}{81.12\%} & 81.78\% & \multicolumn{1}{l|}{70.35\%} & 71.99\% \\ \hline
2          & \multicolumn{1}{l|}{74.95\%} & 76.35\% & \multicolumn{1}{l|}{74.49\%} & 77.13\% & \multicolumn{1}{l|}{87.02\%} & 89.39\% & \multicolumn{1}{l|}{81.35\%} & 81.79\% & \multicolumn{1}{l|}{70.83\%} & 73.35\% \\ \hline
5          & \multicolumn{1}{l|}{75.25\%} & 76.48\% & \multicolumn{1}{l|}{74.50\%} & 77.28\% & \multicolumn{1}{l|}{87.15\%} & 89.58\% & \multicolumn{1}{l|}{81.60\%} & 81.86\% & \multicolumn{1}{l|}{71.01\%} & 73.52\% \\ \hline
8          & \multicolumn{1}{l|}{75.42\%} & 76.57\% & \multicolumn{1}{l|}{75.00\%} & 77.32\% & \multicolumn{1}{l|}{87.17\%} & 89.58\% & \multicolumn{1}{l|}{81.65\%} & 81.94\% & \multicolumn{1}{l|}{71.49\%} & 73.60\% \\ \hline
10         & \multicolumn{1}{l|}{75.50\%} & 76.65\% & \multicolumn{1}{l|}{75.15\%} & 77.40\% & \multicolumn{1}{l|}{89.22\%} & 90.35\% & \multicolumn{1}{l|}{81.66\%} & 81.99\% & \multicolumn{1}{l|}{71.60\%} & 73.70\% \\ \hline
\end{tabular}

\label{epsilons_T} 
\end{table*}

\subsection{Clip vs.  Metric scores}

We analyze the impact of different clip or gradient norm bound $S$ rates (1, 2, 5, 8, and 10) on the model's performance. Table ~\ref{clip_T} presents the results, highlighting the influence of the clip on the model's utility. We observe that the model's performance improves as the clip value increases. For instance, DP achieves precision scores of 72.42\%, 72.75\%, 73.46\%, 73.97\%, and 74.12\% for clips 1, 2, 5, 8, and 10, respectively. In comparison, ADP achieves higher precision scores of 74.09\%, 74.46\%, 75.03\%, 75.24\%, and 75.50\% for the corresponding clip values. This pattern holds for other metrics as well. If the clip value is too low, a significant amount of model information is lost, resulting in poorer performance. ADP consistently outperforms DP across all metrics. For example, at a clip of 10, the accuracy, precision, recall, F1-score, and ROC score for DP are recorded as 74.58\%, 74.12\%, 90.86\%, 81.21\%, and 70.35\%, respectively, while ADP achieves higher scores of 75.78\%, 75.50\%, 91.78\%, 81.84\%, and 71.99\%, respectively. This trend is observed consistently for different clip values. 

\begin{table*}[!hbtp]
\centering
\caption{Clip versus metric scores for DP-Adam and Adaptive DP-AdamW}
\begin{tabular}{|l|ll|ll|ll|ll|ll|}
\hline
clip & \multicolumn{2}{l|}{Accuracy}          & \multicolumn{2}{l|}{Precision}         & \multicolumn{2}{l|}{Recall}            & \multicolumn{2}{l|}{F1 score}          & \multicolumn{2}{l|}{ROC score}         \\ \hline
     & \multicolumn{1}{l|}{DP}      & ADP     & \multicolumn{1}{l|}{DP}      & ADP     & \multicolumn{1}{l|}{DP}      & ADP     & \multicolumn{1}{l|}{DP}      & ADP     & \multicolumn{1}{l|}{DP}      & ADP     \\ \hline
1    & \multicolumn{1}{l|}{73.66\%} & 75.06\% & \multicolumn{1}{l|}{72.42\%} & 74.09\% & \multicolumn{1}{l|}{89.22\%} & 89.58\% & \multicolumn{1}{l|}{80.96\%} & 81.63\% & \multicolumn{1}{l|}{68.58\%} & 70.61\% \\ \hline
2    & \multicolumn{1}{l|}{73.93\%} & 75.29\% & \multicolumn{1}{l|}{72.75\%} & 74.46\% & \multicolumn{1}{l|}{89.55\%} & 89.85\% & \multicolumn{1}{l|}{81.06\%} & 81.71\% & \multicolumn{1}{l|}{68.98\%} & 71.01\% \\ \hline
5    & \multicolumn{1}{l|}{74.38\%} & 75.65\% & \multicolumn{1}{l|}{73.46\%} & 75.03\% & \multicolumn{1}{l|}{90.03\%} & 90.78\% & \multicolumn{1}{l|}{81.12\%} & 81.78\% & \multicolumn{1}{l|}{69.76\%} & 71.61\% \\ \hline
8    & \multicolumn{1}{l|}{74.53\%} & 75.66\% & \multicolumn{1}{l|}{73.97\%} & 75.24\% & \multicolumn{1}{l|}{90.53\%} & 91.52\% & \multicolumn{1}{l|}{81.14\%} & 81.79\% & \multicolumn{1}{l|}{70.22\%} & 71.75\% \\ \hline
10   & \multicolumn{1}{l|}{74.58\%} & 75.78\% & \multicolumn{1}{l|}{74.12\%} & 75.50\% & \multicolumn{1}{l|}{90.86\%} & 91.78\% & \multicolumn{1}{l|}{81.21\%} & 81.84\% & \multicolumn{1}{l|}{70.35\%} & 71.99\% \\ \hline
\end{tabular}
\label{clip_T}
\end{table*}

\subsection{Noise multiplier vs. Metric scores}

We investigate the impact of the noise multiplier (NM) on the model's performance. Table ~\ref{NM_T} presents the performance of both DP and ADP algorithms for five different NM values: 1, 2, 5, 8, and 10. We observe that DP and ADP algorithms experience a decrease in performance as the NM value increases. For example, the recall scores for DP at NM values of 1, 2, 5, 8, and 10 are measured at 89.37\%, 89.23\%, 88.98\%, 88.97\%, and 88.84\%, respectively. Similarly, ADP achieves recall scores of 89.51\%, 89.37\%, 89.37\%, 88.99\%, and 88.98\% for the corresponding NM values. This pattern holds true for other metrics as well. The reason could be increasing the NM increases the model's noise, leading to less accurate results. Moreover, we observe ADP consistently outperforms DP. For instance, at NM of 1, DP achieves accuracy, precision, recall, F1-score, and ROC score of 73.75\%, 73.59\%, 89.37\%, 80.50\%, and 69.51\%, respectively, while ADP achieves higher scores of 74.37\%, 73.95\%, 89.51\%, 80.99\%, and 70.11\%, respectively. Similar patterns are observed for different NM values as well. 

\begin{table*}[!hbtp]
\centering
\caption{Noise multiplier versus metric scores for DP-Adam and Adaptive DP-AdamW}
\begin{tabular}{|l|ll|ll|ll|ll|ll|}
\hline
NM & \multicolumn{2}{l|}{Accuracy}          & \multicolumn{2}{l|}{Precision}         & \multicolumn{2}{l|}{Recall}            & \multicolumn{2}{l|}{F1 score}          & \multicolumn{2}{l|}{ROC score}         \\ \hline
   & \multicolumn{1}{l|}{DP}      & ADP     & \multicolumn{1}{l|}{DP}      & ADP     & \multicolumn{1}{l|}{DP}      & ADP     & \multicolumn{1}{l|}{DP}      & ADP     & \multicolumn{1}{l|}{DP}      & ADP     \\ \hline
1  & \multicolumn{1}{l|}{73.75\%} & 74.37\% & \multicolumn{1}{l|}{73.59\%} & 73.95\% & \multicolumn{1}{l|}{89.37 \%} & 89.51\% & \multicolumn{1}{l|}{80.50\%} & 80.99\% & \multicolumn{1}{l|}{69.51\%} & 70.11\% \\ \hline
2  & \multicolumn{1}{l|}{72.19\%} & 72.29\% & \multicolumn{1}{l|}{71.87\%} & 71.96\% & \multicolumn{1}{l|}{89.23\%} & 89.37\% & \multicolumn{1}{l|}{79.67\%} & 79.73\% & \multicolumn{1}{l|}{67.35\%} & 67.48\% \\ \hline
5  & \multicolumn{1}{l|}{69.92\%} & 70.09\% & \multicolumn{1}{l|}{69.82\%} & 69.94\% & \multicolumn{1}{l|}{88.98\%} & 89.37\% & \multicolumn{1}{l|}{78.35\%} & 78.47\% & \multicolumn{1}{l|}{64.49\%} & 64.67\% \\ \hline
8  & \multicolumn{1}{l|}{68.69\%} & 68.79\% & \multicolumn{1}{l|}{68.81\%} & 68.90\% & \multicolumn{1}{l|}{88.97\%} & 88.99\% & \multicolumn{1}{l|}{77.61\%} & 77.67\% & \multicolumn{1}{l|}{62.98\%} & 63.11\% \\ \hline
10 & \multicolumn{1}{l|}{68.17\%} & 68.29\% & \multicolumn{1}{l|}{68.36\%} & 68.47\% & \multicolumn{1}{l|}{88.84\%} & 88.98\% & \multicolumn{1}{l|}{77.32\%} & 77.38\% & \multicolumn{1}{l|}{62.31\%} & 62.47\% \\ \hline
\end{tabular}
\vspace{-3mm}
\label{NM_T}
\end{table*}
\subsection{Optimizer vs. Metric scores}

We evaluate the performance of different optimizers (SGD, RmsProp, Adam, and AdamW) under the DP and ADP algorithms. From the Table ~\ref{opt_T}, we can see that, AdamW demonstrates the best performance for both DP and ADP. For example, the F1-score values for SGD, RmsProp, Adam, and AdamW optimizers under DP and ADP algorithms are recorded as 78.06\%, 79.09\%, 79.09\%, and 81.12\%, and 79.87\%, 81.40\%, 81.78\%, and 81.79\%, respectively. Similar trends are observed for other metrics as well. Across all the optimizers, ADP consistently outperforms DP. For instance, using the AdamW optimizer, the accuracy, precision, recall, F1-score, and ROC scores under DP and ADP are obtained as follows: (74.58\%, 75.68\%, 77.80\%, 89.22\%, 81.12\%, and 71.98\%), (75.78\%, 77.80\%, 93.07\%, 81.98\%, and 72.28\%). The same pattern is observed  for different optimizers as well.

\begin{table*}[h]
\centering  
\caption{Optimizer versus metric scores for DP-Adam and Adaptive DP-AdamW}
\begin{tabular}{|l|ll|ll|ll|ll|ll|}
\hline
opt & \multicolumn{2}{l|}{Accuracy}          & \multicolumn{2}{l|}{Precision}         & \multicolumn{2}{l|}{Recall}            & \multicolumn{2}{l|}{F1 score}          & \multicolumn{2}{l|}{ROC score}         \\ \hline
\multicolumn{1}{|l|}{}        & \multicolumn{1}{l|}{DP}      & \multicolumn{1}{l|}{ADP}     & \multicolumn{1}{l|}{DP}      & \multicolumn{1}{l|}{ADP}     & \multicolumn{1}{l|}{DP}      & \multicolumn{1}{l|}{ADP}     & \multicolumn{1}{l|}{DP}      & \multicolumn{1}{l|}{ADP}     & \multicolumn{1}{l|}{DP}      & \multicolumn{1}{l|}{ADP}     \\ \hline
\multicolumn{1}{|l|}{SGD}     & \multicolumn{1}{l|}{71.39\%} & \multicolumn{1}{l|}{72.62\%} & \multicolumn{1}{l|}{69.94\%} & \multicolumn{1}{l|}{76.31\%} & \multicolumn{1}{l|}{79.90\%} & \multicolumn{1}{l|}{89.58\%} & \multicolumn{1}{l|}{78.06\%} & \multicolumn{1}{l|}{79.87\%} & \multicolumn{1}{l|}{65.29\%} & \multicolumn{1}{l|}{70.57\%} \\ \hline
\multicolumn{1}{|l|}{RmsProp} & \multicolumn{1}{l|}{74.03\%} & \multicolumn{1}{l|}{75.46\%} & \multicolumn{1}{l|}{74.12\%} & \multicolumn{1}{l|}{77.70\%} & \multicolumn{1}{l|}{80.53\%} & \multicolumn{1}{l|}{88.06\%} & \multicolumn{1}{l|}{79.09\%} & \multicolumn{1}{l|}{81.40\%} & \multicolumn{1}{l|}{71.92\%} & \multicolumn{1}{l|}{72.21\%} \\ \hline
\multicolumn{1}{|l|}{Adam}    & \multicolumn{1}{l|}{74.07\%} & \multicolumn{1}{l|}{75.77\%} & \multicolumn{1}{l|}{75.50\%} & \multicolumn{1}{l|}{75.50\%} & \multicolumn{1}{l|}{80.42\%} & \multicolumn{1}{l|}{89.21\%} & \multicolumn{1}{l|}{79.09\%} & \multicolumn{1}{l|}{81.78\%} & \multicolumn{1}{l|}{70.35\%} & \multicolumn{1}{l|}{71.99\%} \\ \hline
\multicolumn{1}{|l|}{AdamW}   & \multicolumn{1}{l|}{74.58\%} & \multicolumn{1}{l|}{75.78\%} & \multicolumn{1}{l|}{75.68\%} & \multicolumn{1}{l|}{77.80\%} & \multicolumn{1}{l|}{89.22\%} & \multicolumn{1}{l|}{93.07\%} & \multicolumn{1}{l|}{81.12\%} & \multicolumn{1}{l|}{81.79\%} & \multicolumn{1}{l|}{71.98\%} & \multicolumn{1}{l|}{72.28\%} \\ \hline
\end{tabular}
\label{opt_T}
\end{table*}

\subsection{Effect of Weighted Random Sampler and Class weights on metric scores under DP}

To evaluate the effectiveness of techniques employed to address data imbalance in non-private models for tackling the same issue in private models, we incorporate CW and a WRS during the training of the differentially private model. However, upon analyzing the results presented in Table~\ref{Tech_T}, we observe that the WRS and CW techniques are ineffective in resolving the data imbalance problem. Additionally, the utilization of class weighting significantly deteriorates the performance of the differentially private model. For instance, considering the accuracy score, we observe that WRS and CW achieve scores of 73.85\% and 59.93\%, respectively. However, when the DP model is trained without employing any technique to address data imbalance, it achieves an accuracy score of 74.58\%. The poor performance exhibited when using CW and WRS in training the DP model may be attributed to the weights assigned to the class or each sample. This increases noise when penalized for misclassification, rendering the model's predictions ineffective rather than useful.  

\begin{table*}[h]
\centering
\caption{Effect of Weighted Random Sampler and Class weights on metric scores under DP}
\begin{tabular}{|l|ll|ll|ll|ll|ll|}
\hline
Technique & \multicolumn{2}{l|}{Accuracy}          & \multicolumn{2}{l|}{Precision}         & \multicolumn{2}{l|}{Recall}            & \multicolumn{2}{l|}{F1 score}          & \multicolumn{2}{l|}{ROC score}         \\ \hline
          & \multicolumn{1}{l|}{DP}      & ADP     & \multicolumn{1}{l|}{DP}      & ADP     & \multicolumn{1}{l|}{DP}      & ADP     & \multicolumn{1}{l|}{DP}      & ADP     & \multicolumn{1}{l|}{DP}      & ADP     \\ \hline
None      & \multicolumn{1}{l|}{74.58\%} & 75.78\% & \multicolumn{1}{l|}{77.78\%} & 78.65\% & \multicolumn{1}{l|}{89.22\%} & 89.58\% & \multicolumn{1}{l|}{81.12\%} & 81.79\% & \multicolumn{1}{l|}{72.13\%} & 73.54\% \\ \hline
WRS       & \multicolumn{1}{l|}{73.85\%} & 75.34\% & \multicolumn{1}{l|}{74.12\%} & 75.50\% & \multicolumn{1}{l|}{79.96\%} & 81.77\% & \multicolumn{1}{l|}{78.85\%} & 80.18\% & \multicolumn{1}{l|}{70.35\%} & 71.99\% \\ \hline
CW        & \multicolumn{1}{l|}{59.93\%} & 59.66\% & \multicolumn{1}{l|}{62.13\%} & 62.24\% & \multicolumn{1}{l|}{86.05\%} & 87.79\% & \multicolumn{1}{l|}{72.23\%} & 72.77\% & \multicolumn{1}{l|}{52.03\%} & 52.24\% \\ \hline
\end{tabular}
 \vspace{-4mm}
\label{Tech_T}
\end{table*}

\subsection{Effect of number of trainable convolutional layers}
\begin{table*}[h!]
\centering
\caption{Effect of number of trainable convolutional layers on metric scores under DP}
\begin{tabular}{|l|ll|ll|ll|ll|ll|}
\hline
\begin{tabular}[c]{@{}l@{}}\#   trainable \\ conv layers\end{tabular} & \multicolumn{2}{l|}{Accuracy}          & \multicolumn{2}{l|}{Precision}         & \multicolumn{2}{l|}{Recall}            & \multicolumn{2}{l|}{F1 score}          & \multicolumn{2}{l|}{ROC score}         \\ \hline
                                                                      & \multicolumn{1}{l|}{DP}      & ADP     & \multicolumn{1}{l|}{DP}      & ADP     & \multicolumn{1}{l|}{DP}      & ADP     & \multicolumn{1}{l|}{DP}      & ADP     & \multicolumn{1}{l|}{DP}      & ADP     \\ \hline
0                                                                     & \multicolumn{1}{l|}{74.58\%} & 75.78\% & \multicolumn{1}{l|}{74.12\%} & 75.50\% & \multicolumn{1}{l|}{90.77\%} & 89.18\% & \multicolumn{1}{l|}{81.12\%} & 81.79\% & \multicolumn{1}{l|}{72.90\%} & 71.99\% \\ \hline
1                                                                     & \multicolumn{1}{l|}{73.80\%} & 76.48\% & \multicolumn{1}{l|}{72.90\%} & 76.26\% & \multicolumn{1}{l|}{89.58\%} & 89.22\%  & \multicolumn{1}{l|}{80.86\%} & 82.22\% & \multicolumn{1}{l|}{69.03\%} & 81.12\% \\ \hline
\end{tabular}

\label{Trainable_T}
\end{table*}
To examine the influence of the number of trainable convolutional layers on metric scores under DP, we introduce the training of the last convolutional layer along with the classification layer. The results obtained when the DP model is trained with and without the final convolutional layer are presented in Table~\ref{Trainable_T}. It is evident from the table that the performance of ADP is enhanced when training includes the convolutional layer, while the DP model's performance deteriorates when the convolutional layer is trained. For instance, the F1-score of DP decreases from 81.79\% to 80.86\% when trained with the convolutional layer. On the other hand, the F1-score of ADP increases from 81.79\% to 82.22\%. This pattern holds for other metrics as well. The drop in performance for DP can be attributed to the introduction of additional noise resulting from the trainable convolutional layer. In the case of ADP, however, the noise scale decreases during the training process, thereby improving the model's performance even when an additional trainable convolutional layer is introduced.

\subsection{Results comparison}
\begin{table}[h!]
\centering
\caption{Results comparison with existing literature}
\begin{tabular}{|l|l|}
\hline
Authors            & Accuracy                     \\ \hline
~\cite{sai2022modified}       & 89.81\%          \\ \hline
~\cite{zhao2022identification}           & 77.3\%   - 98.7\% \\ \hline
~\cite{li2019early}          & 96.5\%   \\ \hline
~\cite{khan2022transfer}            &  90.91\%                           \\ \hline
Proposed              & 88.98\%                      \\ \hline
Proposed(ADP)         & 76.48\%                     \\ \hline
Proposed(DP)          & 74.58\%                     \\ \hline
\end{tabular}
\vspace{-4mm}
\label{tab16}
\end{table}

From the information presented in Table \ref{tab16}, it is evident that previous studies related to GI cancer did not prioritize the protection of dataset privacy in their respective projects. To the best of our knowledge, we are the first to implement a privacy-enhanced mechanism for classifying GI cancer, specifically for distinguishing between MSI and MSS. In our approach, we utilize two variants of DP to fine-tune the model. However, it is important to note that the accuracy of our non-private model is 88.98\%, while the current state-of-the-art (SOTA) achieves the accuracy of 96.5\%. It is crucial to clarify that the objective of this paper is not to achieve SOTA accuracy using a non-private model. Instead, we focus on leveraging DP to obtain a reasonably accurate model while prioritizing data privacy.

\section{Limitations and Future Work}
A notable performance gap exists between DP and non-private models. This gap can be narrowed by employing more advanced DP mechanisms, such as Dynamic DP-SGD, as proposed by ~\cite{du2021dynamic}. It is worth mentioning that the training time of DP models is significantly longer compared to non-private models. To mitigate this issue, the convolutional operations involved in DP can be accelerated using the techniques described in ~\cite{tida2022kernel}. Additionally, the overall DP procedure can be enhanced by incorporating Johnson-Lindenstrauss (JL) projections, as suggested in ~\cite{bu2021fast}. 

\section{Conclusion}

Classifying the status of cancer tumors as MSI vs. MSS is crucial for providing appropriate treatment to patients while ensuring the privacy of their data. In our study, we focused on fine-tuning the NF-Net model, known for its compatibility with DP, to achieve this task. By fine-tuning the NF-Net F0 feature extractor, we developed a differentially private model capable of predicting the status of GI cancer while safeguarding patient-level information. Our evaluation encompasses various aspects, including the impact of trainable convolutional layers on metric scores under DP, the effects of WRS and CW on metric scores under DP, the relationship between Optimizer and metric scores, the influence of Noise multiplier on metric scores for DP-Adam and Adaptive DP-AdamW, the correlation between Noise multiplier and metric scores, the relationship between clip values and metric scores, and the effects of Epsilon on metric scores. Through our experimentation, we achieved an accuracy of 76.48\% when utilizing an epsilon of 1, one additional trainable convolutional layer in conjunction with the final classifier, and training the model using Adaptive DP-AdamW.

\bibliographystyle{apacite}
\footnotesize
\bibliography{sample}

\end{document}